\documentclass[conference]{IEEEtran}
\IEEEoverridecommandlockouts
\usepackage{booktabs} 
\usepackage{tikz}
\usepackage{forest}
\usetikzlibrary{shapes.geometric, arrows, positioning}

\usepackage{natbib}
\usepackage{amsmath,amssymb,amsfonts}
\usepackage{algorithmic}
\usepackage{graphicx}
\usepackage{textcomp}
\usepackage{xcolor}
\def\BibTeX{{\rm B\kern-.05em{\sc i\kern-.025em b}\kern-.08em
T\kern-.1667em\lower.7ex\hbox{E}\kern-.125emX}}

\usepackage{hyperref}
\hypersetup{
    colorlinks=true,
    linkcolor=darkblue,
    citecolor=darkblue,
    urlcolor=darkblue
}
\definecolor{darkblue}{rgb}{0.0, 0.0, 0.55}

\begin{document}
\title{Mixture of Experts in Large Language Models
\thanks{\hspace*{-\parindent}\rule{3.8cm}{0.4pt} \\ 
$\dagger$: Corresponding author: Junhao Song (junhao.song23@imperial.ac.uk)}
}

\author{
\IEEEauthorblockN{Danyang Zhang}
\IEEEauthorblockA{\textit{Department of Research} \\
\textit{ByteDance Inc}\\
San Jose, USA\\
joseph.zhang@bytedance.com }

\and

\IEEEauthorblockN{Junhao Song$^{\dagger}$}
\IEEEauthorblockA{\textit{Department of CS} \\
\textit{Imperial College London}\\
London, UK \\
junhao.song23@imperial.ac.uk}

\and

\IEEEauthorblockN{Ziqian Bi}
\IEEEauthorblockA{\textit{Department of CS} \\
\textit{Purdue University}\\
Indiana, USA \\
bi32@purdue.edu}

\and

\IEEEauthorblockN{Xinyuan Song}
\IEEEauthorblockA{\textit{Department of CS} \\
\textit{Emory University}\\
Atlanta, USA \\
xinyuan.song@emory.edu}

\and

\IEEEauthorblockN{Yingfang Yuan}
\IEEEauthorblockA{\textit{Department of Computer Science} \\
\textit{Heriot-Watt University}\\
Edinburgh, United Kingdom \\
y.yuan@hw.ac.uk}

\and

\IEEEauthorblockN{Tianyang Wang}
\IEEEauthorblockA{\textit{AI Agent Lab} \\
\textit{Vokram Group}\\
London, United Kingdom \\
tianyang.wg35@gmail.com}

\and

\IEEEauthorblockN{Joe Yeong}
\IEEEauthorblockA{\textit{Department of Anatomical Pathology} \\
\textit{Singapore General Hospital}\\
Singapore \\
yeongps@imcb.a-star.edu.sg}

\and

\IEEEauthorblockN{Junfeng Hao}
\IEEEauthorblockA{\textit{AI Agent Lab} \\
\textit{Vokram Group}\\
London, United Kingdom \\
ygzhjf85@gmail.com}
}

\maketitle

\begin{abstract}
This paper presents a comprehensive review of the Mixture-of-Experts (MoE) architecture in large language models, highlighting its ability to significantly enhance model performance while maintaining minimal computational overhead. Through a systematic analysis spanning theoretical foundations, core architectural designs, and large language model (LLM) applications, we examine expert gating and routing mechanisms, hierarchical and sparse MoE configurations, meta-learning approaches, multimodal and multitask learning scenarios, real-world deployment cases, and recent advances and challenges in deep learning. Our analysis identifies key advantages of MoE, including superior model capacity compared to equivalent Bayesian approaches, improved task-specific performance, and the ability to scale model capacity efficiently. We also underscore the importance of ensuring expert diversity, accurate calibration, and reliable inference aggregation, as these are essential for maximizing the effectiveness of MoE architectures. Finally, this review outlines current research limitations, open challenges, and promising future directions, providing a foundation for continued innovation in MoE architecture and its applications.
\end{abstract}


\begin{figure*}[t]
\centering
\includegraphics[width=\linewidth]{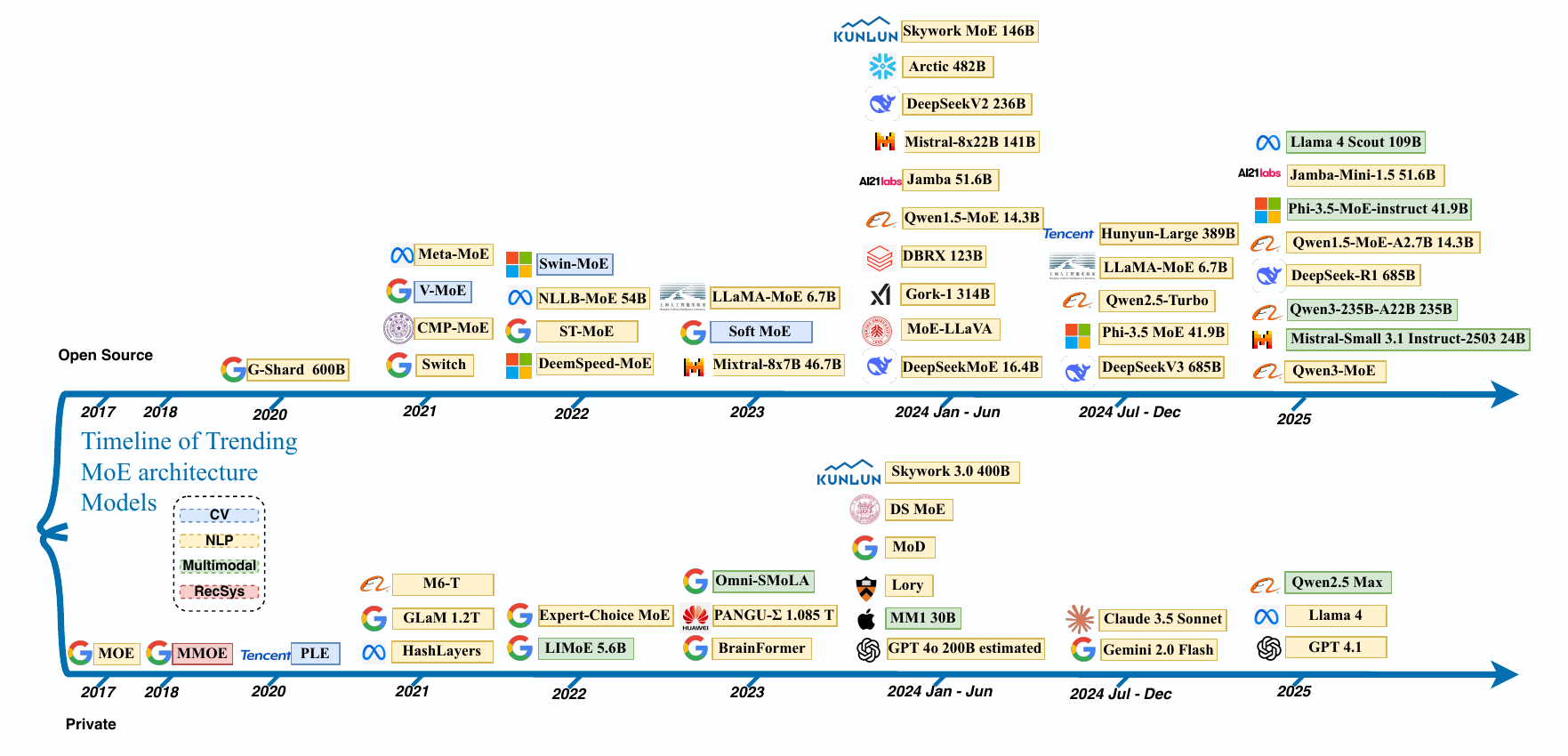}
\caption{\textbf{Timeline of mixture of experts (MoE) models development.} The timeline shows key milestones in MoE architecture evolution from foundational concepts to modern large-scale implementations.}
\label{fig:moe_timeline_teaser}
\end{figure*}

\section{Introduction and Fundamentals}


Over the past decade, deep learning has witnessed an explosive increase in model capacity, particularly with the emergence of large-scale transformer architectures. These models, while powerful, are often computationally intensive and memory hungry. Scaling such systems linearly in parameter count leads to exponential increases in FLOPs and energy consumption. This trend is rapidly becoming unsustainable for deployment in real-world scenarios. In response, researchers have begun exploring sparse and modular computation as viable alternatives to dense, monolithic models. Among these, Mixture-of-Experts (MoE) architectures stand out as a promising solution. By conditionally activating only a subset of a model’s parameters, typically a small number of specialized expert modules, MoEs decouple inference cost from total model size, enabling efficient scaling of model capacity~\cite{shazeer2017outrageously}.

The conceptual foundations of MoE trace back to early work in adaptive learning systems, where models were organized as ensembles of specialized experts, each handling a specific subregion of the input space~\cite{jacobs1991adaptive, xu1994alternative, waterhouse1995bayesian}. These early frameworks emphasized modularity and competitive learning, often using a gating function to route each input to the most relevant expert. However, computational constraints and the lack of scalable training mechanisms limited their practical impact. MoEs did not become viable for high-performance applications until sparse routing was integrated with modern deep networks, made possible by large-scale distributed computation. The breakthrough came with the development of sparsely gated networks~\cite{shazeer2017outrageously}, which showed that it is possible to maintain model accuracy while activating only a small fraction of the parameters during both training and inference.

More than just a tool for parameter efficiency, MoE represents a broader architectural shift toward modular design in deep learning. Traditional networks treat all parts of the model as uniformly important across all tasks and inputs. In contrast, MoEs introduce conditional computation: each expert module can specialize in a certain domain, linguistic pattern, or modality. During training, a learnable gating function, often implemented as a lightweight neural network, determines which experts are best suited to handle each input~\cite{1312.4314, bengio2015conditional}. This routing process encourages functional diversity among the experts, allowing them to focus on distinct aspects of the input distribution, which improves generalization and robustness across tasks.

The renewed interest in MoEs aligns with a growing recognition that monolithic deep learning models, while powerful, are not necessarily optimal from a computational or representational standpoint. MoE architectures offer a more flexible and scalable alternative by embracing heterogeneity in both structure and computation. Their success across diverse application domains, including language modeling, machine translation, and vision-language reasoning, demonstrates the generalizability of the approach. Crucially, MoEs challenge the prevailing assumption that scale must come at the cost of efficiency, providing a viable path toward building more intelligent, adaptable, and resource-conscious AI systems.


\vspace{0.5em}
\textbf{From sparse gating to billion-scale deployment.} The development of MoE architectures underwent a pivotal transformation starting in 2020, when research shifted from conceptual formulations to practical, high-scale implementations. As shown in \textbf{Figure~\ref{fig:moe_timeline_teaser}}, this turning point was marked by the release of GShard~\cite{lepikhingshard}, a large-scale multilingual model with 600 billion parameters. GShard pioneered auto-sharding and token-level expert routing, making it one of the first successful demonstrations of sparse MoE architectures at the trillion-parameter scale. Shortly after, Switch Transformer~\cite{fedus2022switch} and GLaM~\cite{du2022glam} extended this paradigm to language modeling. These models employed token-choice gating with only 1–2 experts activated per input, significantly reducing compute overhead and establishing MoE as a credible scaling alternative to dense transformers.

By 2021–2022, MoE architectures had matured from proof-of-concept models into a broadly adopted computational framework. Open-source releases such as Meta-MoE, CMP-MoE, and V-MoE demonstrated the community-wide interest in modular computation. At the same time, commercial labs began investing in domain-specific variants: NLLB-MoE for multilingual translation, Swin-MoE for vision tasks, and LIMoE~\cite{mustafa2022multimodal} for multimodal learning. This diversification marked the beginning of MoE’s adaptation beyond NLP, driven by the recognition that sparse expert activation could generalize to vision, audio, and cross-modal scenarios.

The post-2023 landscape, as shown in the upper half of \textbf{Figure~\ref{fig:moe_timeline_teaser}}, marks a phase of rapid industrial scaling and architectural diversification. Models such as DeepSeekV3 (685B), Skywork 3.0 (400B), and Arctic (482B) demonstrate that MoE has become a core component of modern foundation models. These systems increasingly integrate expert routing with paradigms like retrieval, instruction tuning, and agent-based control, indicating that MoE now plays a central role in large-scale AI development. The emergence of models such as MoE-LLaVA, MM1~\cite{mckinzie2024mm1}, and Omni-SMoLA~\cite{wu2024omni} further reflects a shift toward multimodal and grounded reasoning architectures.

Open-source initiatives have accelerated MoE adoption through the release of advanced models such as Jamba, Qwen1.5-MoE, and Mistral-8x22B, while continued improvements to routing and load balancing in Qwen3-MoE, Claude~3.5~Sonnet, and Llama~4 indicate that innovation now centers on efficiency, controllability, and multi-task generalization rather than scale alone. From mid-2025 onward, the main technical focus further shifts from increasing parameter count to making routing reliable under long training runs and deployment constraints. On the routing side, Omi et al.~\cite{omi2025similarity_router} propose a similarity-preserving load-balancing objective that stabilizes expert selection for related inputs while avoiding expert collapse, enabling faster and less redundant training. Complementarily, Dong et al.~\cite{dong2025maxscore} formulate routing as a constrained optimization problem via \textsc{MaxScore}, addressing token dropping and padding inefficiency caused by hard expert-capacity constraints. At the survey level, Zhang et al.~\cite{zhang2025moe_review} summarize mid-2025 MoE design choices, including gating strategies, sparse and hierarchical variants, multimodal extensions, and deployment considerations, showing that evaluation increasingly emphasizes expert diversity, calibration, and inference aggregation. In late 2025, Qwen3-VL~\cite{bai2025qwen3vl} reports vision--language MoE variants (for example, 30B-A3B and 235B-A22B) designed for latency--quality trade-offs with long interleaved multimodal context, indicating that MoE is now treated as a practical control for compute allocation across modalities rather than a scaling-only mechanism.

\begin{table*}[t]
\centering
\renewcommand\arraystretch{0.8}
\caption{Taxonomy of representative MoE architectures across six application domains, highlighting expert counts, routing strategies, development periods (2017–2024), and key innovations or use cases.}
\resizebox{\linewidth}{!}{%
\begin{tabular}{@{}llccc p{5.0cm}@{}}
\toprule[1pt]
\textbf{Category} & \textbf{Model (\#Experts)} & \textbf{Routing} & \textbf{Year} & \textbf{Key Innovation / Use Case} \\ \midrule
Language LLM & Switch Transformer (64)~\cite{fedus2022switch} / GLaM (64)~\cite{du2022glam} & Token-choice & 2021–22 & Trillion-parameter LLMs with only 1/64 active parameters per token. \\
Translation  & GShard MoE (128)~\cite{lepikhingshard} / DeepSpeed-MoE (256)~\cite{rajbhandari2022deepspeed} & Token-choice & 2020–21 & Auto sharding + pipeline parallelism for cross-lingual MT at TB scale. \\
Multimodal   & Omni-SMoLA (16)~\cite{2312.00968} / T-REX2 (32)~\cite{jiang2024t} & Cross-modal gate & 2023–24 & Low-rank experts plus dual vision–text prompts for open-set detection. \\
Computer Vision & MoCaE-DET (8)~\cite{2309.14976} / Deep-MoE (32)~\cite{1312.4314} & Attention gate & 2017–23 & Calibration-aware fusion; boosts COCO AP by 2.5 vs. single detector. \\
Param-Efficient & LoRA-MoE (4)~\cite{2411.18466} / Nexus (8)~\cite{2408.15901} & Frozen router & 2023–24 & $<$1 \% parameter update (PEFT); “upcycles” dense checkpoints to adaptive MoE. \\
Hierarchical & H-MoE (32)~\cite{2410.02935} / MixER (10)~\cite{2502.05335} & 2-level / Top-1 & 2024 & Coarse-to-fine quadratic gating; K-means routing for dynamical systems. \\ 
\bottomrule[1pt]
\end{tabular}}
\label{table:taxonomy}
\end{table*}

\vspace{0.5em}
\textbf{Diversity of designs across modalities and tasks.} The MoE paradigm has evolved into a diverse architectural space, as shown in Table~\ref{table:taxonomy}. Its variants differ not only in scale and number of experts but also in routing strategies and domain-specific design choices. Early implementations such as Switch Transformer and GLaM~\cite{fedus2022switch, du2022glam} focused on language modeling, where token-level gating activates only a subset of experts per token. This form of sparse activation enabled scaling to hundreds of billions of parameters with minimal compute overhead, while preserving sequence-level modeling fidelity.

In contrast, translation-oriented models like GShard and DeepSpeed-MoE~\cite{lepikhingshard, rajbhandari2022deepspeed} emphasize system-level throughput. These models combine MoE with pipeline parallelism and expert sharding to process multilingual data at terabyte scale. Multimodal variants including Omni-SMoLA~\cite{2312.00968} and T-REX2~\cite{jiang2024t} route using both visual and textual cues, enabling open-set recognition and grounded captioning where context spans modalities.

Other architectures diverge along structural dimensions. Hierarchical models such as H-MoE~\cite{2410.02935} and MixER~\cite{2502.05335} implement multi-stage routing, often using clustering or coarse-to-fine attention to improve interpretability and modularity. Parameter-efficient frameworks like LoRA-MoE~\cite{2411.18466} and Nexus~\cite{2408.15901} focus on minimizing update costs during fine-tuning, by freezing the router and only adapting a few expert weights. These designs are especially useful in settings where retraining large models is computationally prohibitive.

\vspace{0.5em}
\textbf{Routing complexity and the stability efficiency tradeoff.}  
The routing algorithm lies at the heart of every MoE system.
Early designs employed simple top-$k$ selection using softmax over expert logits~\cite{shazeer2017outrageously}, but recent variants increasingly explore entropy-based~\cite{zhou2022mixture}, load-balanced~\cite{fedus2022switch}, or differentiable attention-based routing mechanisms~\cite{lewis2021base, zhou2022mixture, riquelme2021scaling}. The goal is twofold: ensure expert diversity while minimizing redundant computation.

Yet, these methods often suffer from instability or underutilization. For example, overly confident gates can collapse to a few dominant experts, whereas uniform allocation reduces specialization. Techniques like auxiliary load balancing losses or stochastic gating~\cite{fedus2022switch} have been proposed, but they introduce additional hyperparameters and training complexity. Designing robust, general-purpose routing remains one of the most technically challenging aspects of MoE development.

\vspace{0.5em}
\textbf{Deployment constraints.} Despite their computational advantages, MoE models face non-trivial barriers to deployment. Sparse expert activation introduces irregular memory access patterns and frequent cross-device communication, resulting in elevated inference latency and hardware underutilization. Moreover, the stochastic nature of routing leads to unstable batching, fragmented workloads, and poor reproducibility. These issues are particularly problematic in low-latency or memory-constrained environments.

To mitigate these issues, newer designs adopt more deployment-aware strategies. LoRA-MoE~\cite{2411.18466} and Nexus~\cite{2408.15901}, for example, freeze routing weights or apply low-rank adapters to stabilize expert usage during inference. These methods reduce routing variance, simplify caching, and enable efficient fine-tuning. This marks a shift from MoE systems optimized purely for scale toward architectures designed with practical usability and system compatibility in mind.

Recent work places stronger emphasis on deployment-aware evaluation and design. A comprehensive survey by Zhang et al.~\cite{zhang2025moe_review} highlights that modern assessment of MoE systems increasingly focuses on expert utilization balance, calibration, and inference-time aggregation behavior rather than parameter count. In parallel, systems such as Qwen3-VL~\cite{bai2025qwen3vl} demonstrate that routing policies can be shaped to meet practical latency and memory constraints in multimodal inference. Together, these results indicate a shift from MoE architectures optimized mainly for training scale toward designs shaped by robustness, efficiency, and deployment feasibility.

\vspace{0.5em}
\textbf{Scope of this work.}  
This paper presents a structured synthesis of recent advances in MoE architectures. It begins by analyzing core components such as expert modules, gating mechanisms, and load balancing strategies. The discussion then extends to domain-specific adaptations in natural language processing, computer vision, and multimodal learning. Key technical challenges are examined, including routing instability, expert underutilization, and scalability limits. Finally, the survey highlights emerging directions such as sparse mixture fusion, expert replay, and lifelong modularity. By consolidating architectural innovations and comparing their design principles, this work clarifies the current landscape and outlines future paths for developing scalable and efficient MoE-based AI systems.

\definecolor{softblue}{RGB}{240, 248, 255}      
\definecolor{mintgreen}{RGB}{240, 255, 240}     
\definecolor{warmgray}{RGB}{255, 245, 238}      
\definecolor{lavender}{RGB}{248, 240, 255}      
\definecolor{peach}{RGB}{255, 248, 220}         
\definecolor{sage}{RGB}{245, 255, 250}          
\definecolor{dustyrose}{RGB}{255, 240, 245}     

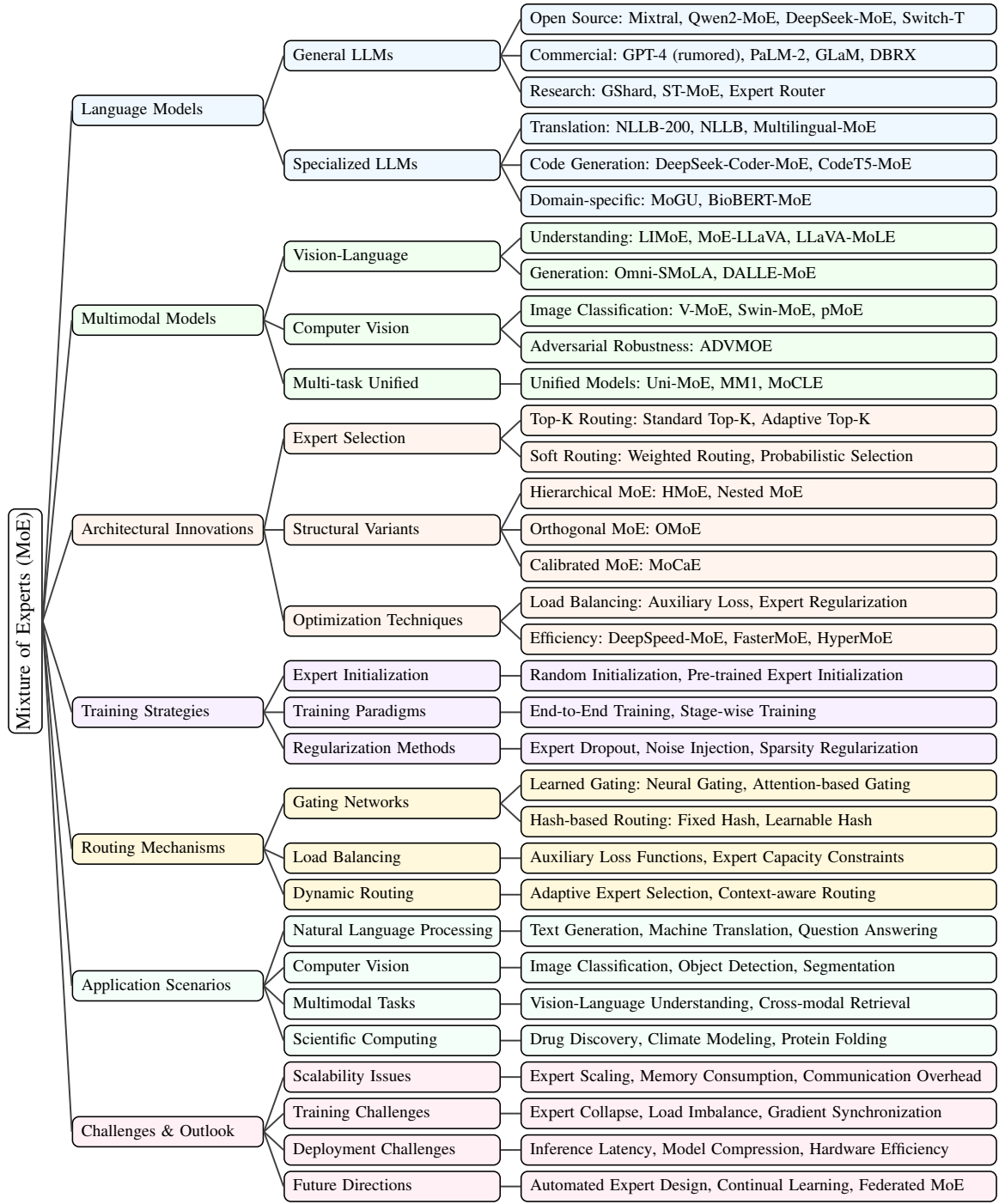
\begin{figure*}[htbp]
    \vspace{-2mm}
    \centering
    \resizebox{0.85\textwidth}{!}{
        \begin{forest}
            for tree={
                child anchor=west,
                parent anchor=east,
                grow'=east,
                anchor=west,
                base=left,
                font=\large,
                rectangle,
                draw=black,
                rounded corners,
                align=left,
                minimum width=4em,
                edge+={darkgray, line width=1pt},
                s sep=3pt,
                inner xsep=2pt,
                inner ysep=3pt,
                line width=0.8pt,
                ver/.style={rotate=90, child anchor=north, parent anchor=south, anchor=center},
language/.style={fill=softblue},
multimodal/.style={fill=mintgreen},
architecture/.style={fill=warmgray},
training/.style={fill=lavender},
routing/.style={fill=peach},
application/.style={fill=sage},
challenge/.style={fill=dustyrose},
            },
            where level=1{text width=11em,font=\normalsize,}{},
            where level=2{text width=12.5em,font=\normalsize,}{},
            where level=3{text width=12em,font=\normalsize,}{},
            where level=4{text width=17em,font=\normalsize,}{},
            [
                Mixture of Experts (MoE), ver 
                [
                    ~Language Models, language
                    [
                        ~General LLMs, language
                       [
    ~Open Source{:} Mixtral{,} Qwen2-MoE{,} DeepSeek-MoE{,} Switch-T
    , language, text width=28em 
]
                        [
                            ~Commercial{:} GPT-4 (rumored){,} PaLM-2{,} GLaM{,} DBRX
                            , language, text width=28em 
                        ]
                        [
                            ~Research{:} GShard{,} ST-MoE{,} Expert Router
                            , language, text width=28em 
                        ]
                    ]
                    [
                        ~Specialized LLMs, language
                        [
                            ~Translation{:} NLLB-200{,} NLLB{,} Multilingual-MoE
                            , language, text width=28em 
                        ]
                        [
                            ~Code Generation{:} DeepSeek-Coder-MoE{,} CodeT5-MoE
                            , language, text width=28em 
                        ]
                        [
                            ~Domain-specific{:} MoGU{,} BioBERT-MoE
                            , language, text width=28em 
                        ]
                    ]
                ]
                [
                    ~Multimodal Models, multimodal
                    [
                        ~Vision-Language, multimodal
                        [
                            ~Understanding{:} LIMoE{,} MoE-LLaVA{,} LLaVA-MoLE
                            , multimodal, text width=28em 
                        ]
                        [
                            ~Generation{:} Omni-SMoLA{,} DALLE-MoE
                            , multimodal, text width=28em 
                        ]
                    ]
                    [
                        ~Computer Vision, multimodal
                        [
                            ~Image Classification{:} V-MoE{,} Swin-MoE{,} pMoE
                            , multimodal, text width=28em 
                        ]
                        [
                            ~Adversarial Robustness{:} ADVMOE
                            , multimodal, text width=28em 
                        ]
                    ]
                    [
                        ~Multi-task Unified, multimodal
                        [
                            ~Unified Models{:} Uni-MoE{,} MM1{,} MoCLE
                            , multimodal, text width=28em 
                        ]
                    ]
                ]
                [
                    ~Architectural Innovations, architecture
                    [
                        ~Expert Selection, architecture
                        [
                            ~Top-K Routing{:} Standard Top-K{,} Adaptive Top-K
                            , architecture, text width=28em 
                        ]
                        [
                            ~Soft Routing{:} Weighted Routing{,} Probabilistic Selection
                            , architecture, text width=28em 
                        ]
                    ]
                    [
                        ~Structural Variants, architecture
                        [
                            ~Hierarchical MoE{:} HMoE{,} Nested MoE
                            , architecture, text width=28em 
                        ]
                        [
                            ~Orthogonal MoE{:} OMoE
                            , architecture, text width=28em 
                        ]
                        [
                            ~Calibrated MoE{:} MoCaE
                            , architecture, text width=28em 
                        ]
                    ]
                    [
                        ~Optimization Techniques, architecture
                        [
                            ~Load Balancing{:} Auxiliary Loss{,} Expert Regularization
                            , architecture, text width=28em 
                        ]
                        [
                            ~Efficiency{:} DeepSpeed-MoE{,} FasterMoE{,} HyperMoE
                            , architecture, text width=28em 
                        ]
                    ]
                ]
                [
                    ~Training Strategies, training
                    [
                        ~Expert Initialization, training
                        [
                            ~Random Initialization{,} Pre-trained Expert Initialization
                            , training, text width=28em 
                        ]
                    ]
                    [
                        ~Training Paradigms, training
                        [
                            ~End-to-End Training{,} Stage-wise Training
                            , training, text width=28em 
                        ]
                    ]
                    [
                        ~Regularization Methods, training
                        [
                            ~Expert Dropout{,} Noise Injection{,} Sparsity Regularization
                            , training, text width=28em 
                        ]
                    ]
                ]
                [
                    ~Routing Mechanisms, routing
                    [
                        ~Gating Networks, routing
                        [
                            ~Learned Gating{:} Neural Gating{,} Attention-based Gating
                            , routing, text width=28em 
                        ]
                        [
                            ~Hash-based Routing{:} Fixed Hash{,} Learnable Hash
                            , routing, text width=28em 
                        ]
                    ]
                    [
                        ~Load Balancing, routing
                        [
                            ~Auxiliary Loss Functions{,} Expert Capacity Constraints
                            , routing, text width=28em 
                        ]
                    ]
                    [
                        ~Dynamic Routing, routing
                        [
                            ~Adaptive Expert Selection{,} Context-aware Routing
                            , routing, text width=28em 
                        ]
                    ]
                ]
                [
                    ~Application Scenarios, application
                    [
                        ~Natural Language Processing, application
                        [
                            ~Text Generation{,} Machine Translation{,} Question Answering
                            , application, text width=28em 
                        ]
                    ]
                    [
                        ~Computer Vision, application
                        [
                            ~Image Classification{,} Object Detection{,} Segmentation
                            , application, text width=28em 
                        ]
                    ]
                    [
                        ~Multimodal Tasks, application
                        [
                            ~Vision-Language Understanding{,} Cross-modal Retrieval
                            , application, text width=28em 
                        ]
                    ]
                    [
                        ~Scientific Computing, application
                        [
                            ~Drug Discovery{,} Climate Modeling{,} Protein Folding
                            , application, text width=28em 
                        ]
                    ]
                ]
                [
                    ~Challenges \& Outlook, challenge
                    [
                        ~Scalability Issues, challenge
                        [
                            ~Expert Scaling{,} Memory Consumption{,} Communication Overhead
                            , challenge, text width=28em 
                        ]
                    ]
                    [
                        ~Training Challenges, challenge
                        [
                            ~Expert Collapse{,} Load Imbalance{,} Gradient Synchronization
                            , challenge, text width=28em 
                        ]
                    ]
                    [
                        ~Deployment Challenges, challenge
                        [
                            ~Inference Latency{,} Model Compression{,} Hardware Efficiency
                            , challenge, text width=28em 
                        ]
                    ]
                    [
                        ~Future Directions, challenge
                        [
                            ~Automated Expert Design{,} Continual Learning{,} Federated MoE
                            , challenge, text width=28em 
                        ]
                    ]
                ]
            ]
        \end{forest}
    }
   \caption{A comprehensive taxonomy of Mixture of Experts (MoE) models, organizing methodologies into seven key categories: language models, multimodal models, architectural innovations, training strategies, routing mechanisms, application scenarios, and challenges. Each category encompasses specific techniques, implementations, and representative models.}
    \label{fig:moe_taxonomy}
\end{figure*}

\section{Core Architectures and Routing Mechanisms}
This section examines the foundational and advanced architectural components of MoE models, with a particular focus on gating mechanisms, expert routing strategies, and architectural variations that promote specialization and efficiency. To provide a structured overview of these developments, \textbf{Figure~\ref{fig:moe_taxonomy}} presents a comprehensive taxonomy of MoE systems. The taxonomy covers seven dimensions, mapping techniques, implementations, and representative systems across language, multimodal, routing, and deployment.

\textbf{Core concepts and mathematical principles.}  
A MoE layer routes each input $x$ to a sparse subset of $k$ expert networks among a larger pool of $N$ experts. Instead of processing the input through all experts, the model computes the output as a weighted combination of only the selected ones:
\begin{equation}
    y = \sum_{i=1}^{N} g_i(x) E_i(x),
\end{equation}
where $E_i(x)$ denotes the output of expert $i$, and $g_i(x)$ is the gating function that assigns nonzero weights to at most $k \ll N$ experts. The gating function is typically implemented via Noisy Top-$k$ routing~\cite{shazeer2017outrageously}, which introduces Gaussian noise into the expert scores before selecting the top $k$:
\begin{equation}
    H(x)_i = (x \cdot W_g)_i + \mathcal{N}(0, \sigma^2)
\end{equation}
This stochasticity prevents early expert collapse and encourages exploration, while top-$k$ sparsity ensures linear compute cost with respect to the number of selected experts, not total expert pool size. The resulting sparse activations can be efficiently parallelized across devices, enabling scalable training and inference.

\subsection{Foundational MoE Architectures}
\begin{figure*}[t]
\centering
\includegraphics[width=\linewidth]{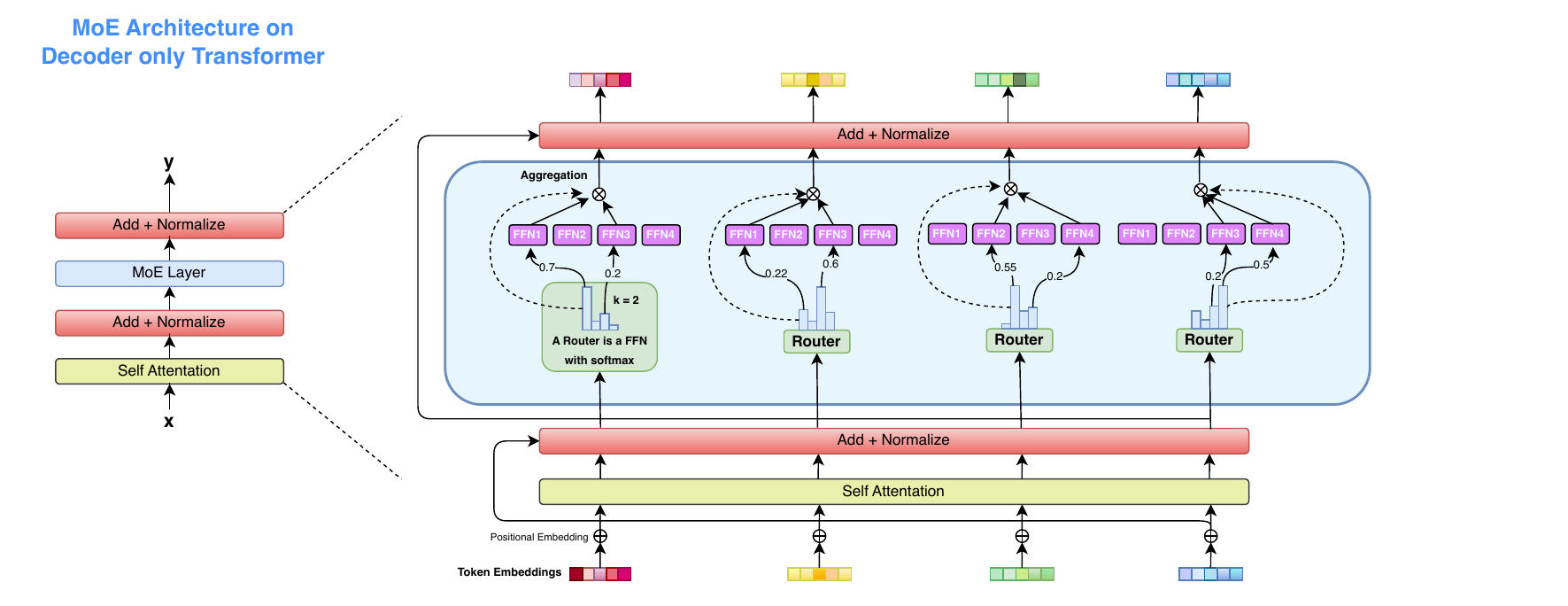}
\caption{\textbf{A brief illustration of sparsely gated Mixture of Experts (MoE) architecture on decoder only transformer.} In this figure, the top-$k$ routing mechanism is configured with $k$=2, meaning the gating function selects the two highest-scoring FFN experts for each token based on the router's softmax probabilities. The selected experts are evaluated in parallel, with their outputs aggregated using weighted summation.}
\label{fig:moe_architecture}
\end{figure*}

\textbf{Sparse activation via gating networks.}  
Each MoE layer comprises $N$ experts $\{E_1, \dots, E_N\}$ and a gating function $G(x)$ that selects a sparse subset of $k \ll N$ experts for each input. As illustrated in \textbf{Figure~\ref{fig:moe_architecture}}, these experts are typically independent feedforward modules. The gating function computes a relevance score $H(x)_i$ for each expert, perturbed by Gaussian noise scaled by a soft activation term:
\begin{align}
    H(x)_i &= (x \cdot W_g)_i + \mathcal{N}(0,1) \cdot \text{Softplus}((x \cdot W_n)_i), \\
    G(x) &= \text{Softmax}(\text{TopK}(H(x), k)),
\end{align}
where $\text{TopK}$ masks all but the $k$ largest scores with $-\infty$ before applying softmax. This Noisy Top-$k$ routing mechanism~\cite{shazeer2017outrageously} promotes exploration, mitigates early expert collapse, and ensures sparse yet balanced expert activation.

By evaluating only a small subset of experts per token, MoE models achieve substantial computational savings and parallel scalability. Large-scale implementations such as Switch Transformer~\cite{fedus2022switch}, GShard~\cite{lepikhingshard}, and DeepSpeed-MoE~\cite{rajbhandari2022deepspeed} demonstrate that this design enables efficient training of models with hundreds of billions of parameters.

\textbf{Load balancing objectives.}  
A key challenge in MoE training is expert collapse—where only a small subset of experts receive the majority of inputs, leaving others underutilized. To mitigate this, a load-balancing objective is added to the training loss:
\begin{equation}
    L_{\text{balance}} = \alpha \sum_{i=1}^{N} f_i \cdot P_i,
\end{equation}
where $f_i$ denotes the fraction of tokens assigned to expert $i$, and $P_i$ is its average gate probability~\cite{fedus2022switch}. The product $f_i \cdot P_i$ penalizes deviations between expected and actual expert usage, encouraging uniform load distribution.

This auxiliary loss promotes better resource utilization and prevents training instability due to expert starvation. However, it introduces a tension with the primary routing objective: enforcing balance may degrade expert specialization or routing accuracy. As a result, selecting the appropriate $\alpha$ and balancing regularization strength remains an open trade-off in large-scale MoE design.

\textbf{From prototypes to production.}  
Transitioning MoE models from conceptual sparsity mechanisms to production-grade systems required addressing significant engineering constraints—particularly in communication overhead, memory fragmentation, and inference latency. Early models like GShard~\cite{lepikhingshard} demonstrated the feasibility of expert parallelism via auto-sharded tensor computation, but required substantial infrastructure support.

Recent advances focus on improving throughput and stability during both training and deployment. For instance, Mixtral~\cite{jiang2024mixtral} employs static top-2 routing and fused attention layers to minimize communication overhead. DBRX~\cite{gupta2024dbrx} introduces fused MoE kernels and low-overhead memory prefetching, while Qwen2~\cite{team2024qwen2} and DeepSeek-v3~\cite{liu2024deepseek} integrate quantized MoE layers and expert dropout to reduce inference cost without degrading accuracy.

These systems increasingly adopt fixed-capacity experts and static load balancing schemes, making them compatible with modern serving frameworks. This convergence of sparsity, hardware alignment, and inference regularization marks MoE's shift from research prototype to scalable, production-ready backbone.

\textbf{Theoretical capacity and scaling behavior.}  
MoE models benefit from a composite hypothesis space defined by a mixture of subnetworks. Let $\mathcal{H}_i$ denote the hypothesis class of expert $E_i$, and $\mathcal{G}$ the gating space. Then the overall capacity satisfies:
\begin{equation}
\mathcal{H}_{\text{MoE}} = \bigcup_{g \in \mathcal{G}} \left\{ \sum_{i=1}^{N} g_i(x) E_i(x) \,\middle|\, E_i \in \mathcal{H}_i \right\}
\end{equation}
where $\mathcal{H}_i$ is the hypothesis class of expert $E_i$, and $\mathcal{G}$ denotes the space of all gating functions.

While MoEs offer high capacity through modular specialization, their lack of shared inductive bias can hinder generalization when routing is unstable. Training dynamics are sensitive to expert overlap and gate smoothness, and unlike Bayesian ensembles, MoEs delegate via hard routing rather than aggregate over uncertainty—potentially increasing variance. Nonetheless, large-scale evaluations show that MoEs achieve comparable performance to dense models on language tasks while activating $\sim$10$\times$ fewer parameters per token~\cite{2501.16352}, demonstrating an effective trade-off between compute efficiency and representational specialization.

\subsection{Advanced Architectural Variants}

\textbf{Orthogonal training.}  
To reduce redundancy and encourage specialization among experts, Orthogonal MoE (OMoE) introduces a regularization term that enforces pairwise weight orthogonality:
\begin{equation}
    L_{\text{orth}} = \sum_{i \neq j} \langle W_i, W_j \rangle^2,
\end{equation}
penalizing alignment between expert parameters and promoting functional diversity~\cite{2501.10062}.

Alternatively, mutual distillation encourages knowledge sharing by minimizing pairwise KL divergence across expert outputs:
\begin{equation}
    L_{\text{distill}} = \sum_{i=1}^{N} \sum_{j \neq i} \text{KL}(E_i(x) \| E_j(x)),
\end{equation}
which improves robustness but risks homogenization if not properly balanced~\cite{2402.00893}.

\textbf{Parameter efficient tuning.}  
MoE architectures can be tuned with high efficiency by restricting updates to lightweight expert-specific components. This allows models to maintain the benefits of sparse computation while drastically reducing the number of trainable parameters. Formally, the parameter update ratio $\rho$ is minimized:
\begin{equation}
    \rho = \frac{\|\Delta \theta_{\text{train}}\|_0}{\|\theta_{\text{full}}\|_0} \ll 1,
\end{equation}
where $\Delta \theta_{\text{train}}$ denotes updated parameters and $\theta_{\text{full}}$ is the total parameter set. Methods such as MoCE-IR~\cite{2411.18466} and Adamix~\cite{wang2022adamix} implement this by freezing shared layers and updating only expert heads or low-rank adapters, often achieving $<1\%$ parameter updates with minimal performance drop.

These strategies align with the goals of Low-Rank Adaptation (LoRA) and adapter-based fine-tuning, treating MoE as a modular scaffold that enables expressive adaptation under strict compute budgets.

\vspace{0.5em}
\textbf{Hierarchical and multi-head extensions.}  
To further enhance specialization and routing flexibility, hierarchical MoEs (HMoEs) introduce a two-stage gating process~\cite{2410.02935}. A coarse gate $G^{(1)}(x)$ selects a super-expert group, while a secondary gate $G^{(2)}(x)$ routes the input within that group:
\begin{equation}
    y = \sum_{i \in \mathcal{G}^{(1)}(x)} G^{(2)}_i(x) E_i(x),
\end{equation}
where $\mathcal{G}^{(1)}(x)$ denotes the selected coarse cluster. This structure enables specialization at multiple abstraction levels and supports large expert pools without routing overhead explosion.

Multi-head MoEs~\cite{2404.15045}, in contrast, assign separate expert subsets to different input dimensions or tasks, allowing input features to be processed in parallel across heads. This setup has shown benefits in vision and speech domains where spatial or temporal decompositions are naturally aligned with the modular structure of MoE.

\textbf{Heterogeneous and adaptive experts.}  
Traditional MoEs assume homogeneous experts with uniform capacity and architecture, but recent advances explore heterogeneity to improve routing flexibility and efficiency~\cite{2408.10681}. In this setup, each expert $E_i$ is characterized by its own computational profile $\phi_i$ (e.g., depth, width, modality), allowing the gating function to assign complex inputs to high-capacity experts and simpler ones to lightweight modules:
\begin{equation}
    g(x) = \arg\max_{i} \left[ S_i(x) - \lambda \cdot \text{Cost}(\phi_i) \right],
\end{equation}
where $S_i(x)$ is the score assigned to expert $i$ and $\text{Cost}(\phi_i)$ reflects its computational expense. This cost-aware routing improves both specialization and hardware efficiency.

\begin{figure*}[t]
\centering
\includegraphics[width=\linewidth]{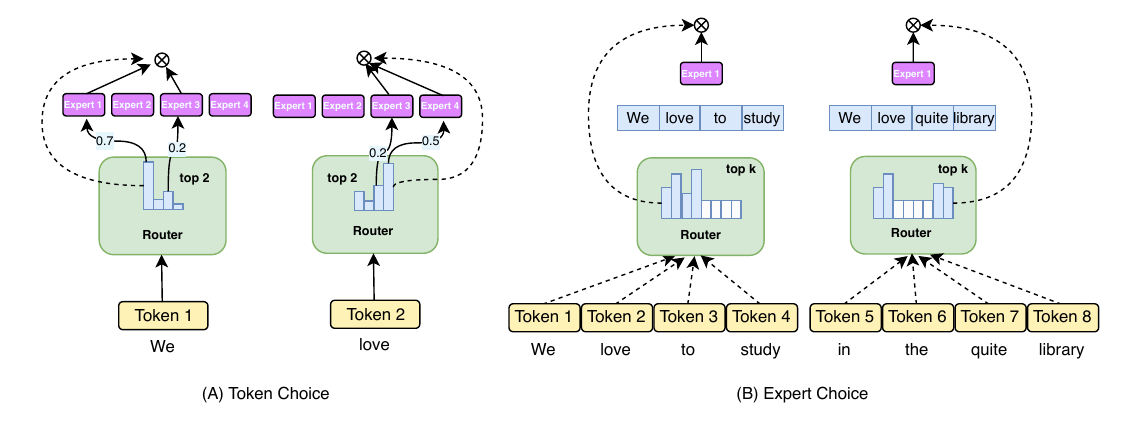}
\caption{\textbf{Comparison of routing strategies in Token Choice and Expert Choice Mixture-of-Experts architectures. } (A) Token Choice routing: Each token is processed by selecting the most suitable experts based on computed affinity scores, with tokens "We" being routed to Expert 1 and Expert 3, while "Like" being routed to and Expert 3 and Expert4, respectively with their corresponding probability weights. (B) Expert Choice routing: Experts maintain fixed computational budgets and select their preferred tokens from the input sequence, where Expert 1 processes tokens ["We", "Love", "To", "Study"] and Expert 2 handles ["We", "Love", "Quite", "Library"], enabling balanced workload distribution across experts while allowing tokens to be processed by multiple experts when beneficial.}
\label{fig:token_choice_expert_choice}
\end{figure*}

Moreover, knowledge integration from unselected experts has emerged as a mechanism for expanding capacity without increasing runtime cost. HyperMoE~\cite{2402.12656}, for instance, aggregates intermediate signals from unselected experts to refine final predictions:
\begin{equation}
    y = \sum_{i \in \mathcal{A}(x)} g_i(x) E_i(x) + \gamma \sum_{j \notin \mathcal{A}(x)} h_j(x),
\end{equation}
where $\mathcal{A}(x)$ is the active expert set, and $h_j(x)$ encodes side information from inactive experts. This promotes cross-task generalization and has shown particular promise in multitask or low-resource regimes.

\subsection{Routing Strategies and Specialization Patterns}

\textbf{Token choice vs. expert choice.}  
Sparse routing in MoE layers can follow either \emph{Token Choice} or \emph{Expert Choice}, as shown in \textbf{Figure~\ref{fig:token_choice_expert_choice}}. In Token Choice, each token $x_t$ is routed to its top-$k$ experts:
\begin{equation}
    y_t = \sum_{i \in \mathcal{A}_t} g_i(x_t) E_i(x_t), \quad \mathcal{A}_t = \text{TopK}_{i=1}^N \left[ g_i(x_t) \right].
\end{equation}
For example, in the left panel, token \texttt{“We”} selects experts 1 and 3, while \texttt{“love”} selects experts 2 and 4, each independently routed based on its own gate scores.

In contrast, Expert Choice lets each expert $E_i$ selects a subset of tokens $\mathcal{T}_i$ to process under a fixed budget:
\begin{equation}
    y_t = \sum_{i: x_t \in \mathcal{T}_i} \tilde{g}_i(x_t) E_i(x_t), \quad \mathcal{T}_i = \text{TopB}_{t=1}^T \left[ s_i(x_t) \right].
\end{equation}
In the right panel, Expert 1 selects tokens \texttt{“We”}, \texttt{“love”}, \texttt{“to”}, and \texttt{“study”}, while Expert 2 handles \texttt{“quiet”} and \texttt{“library”}, enabling token grouping and controlled load balancing.

These paradigms differ fundamentally in control flow: Token Choice assigns experts per-token, while Expert Choice assigns tokens per-expert. The latter improves expert utilization and coherence, particularly in vision or structured input tasks~\cite{2401.15969}.

\begin{figure*}[t]
\centering
\includegraphics[width=\linewidth]{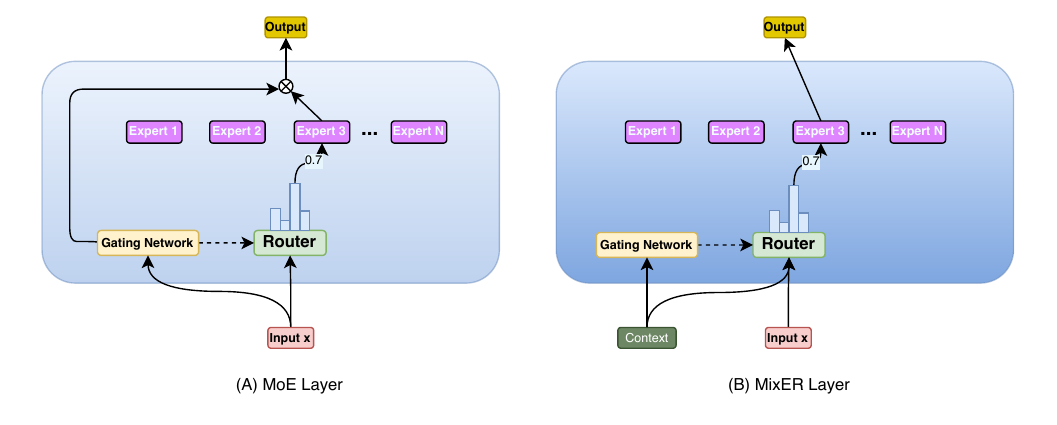}
\caption{\textbf{Architectural comparison between standard MoE and MixER layers.} (A) Standard MoE architecture where input $x$ flows through the gating network to generate routing decisions, directing computation to selected experts via the router mechanism. (B) Enhanced MixER layer design that incorporates an additional context vector $\xi$ alongside input x for routing decisions. The gating network leverages both inputs to compute expert selection probabilities, while the MixER approach eliminates the traditional softmax-weighted output combination used in conventional MoE implementations}
\label{fig:MixER}
\end{figure*}
\textbf{Learned routers vs. Fixed routers.}  
Although trainable gating networks are widely used, recent work~\cite{2402.13089} shows that randomly initialized fixed routers can yield comparable or even superior performance. This challenges the assumption that learned gating $g_\theta(x)$ necessarily improves specialization. In fixed routing, the gate $g(x)$ becomes a deterministic mapping or a sampled sparse mask:
\begin{equation}
    g_i(x) = \begin{cases}
    1, & i \in \mathcal{A}_{\text{fixed}}(x) \\
    0, & \text{otherwise}
    \end{cases}
\end{equation}
where $\mathcal{A}_{\text{fixed}}(x)$ is a static assignment determined at initialization. This approach avoids routing instability and eliminates variance from gradient updates, especially in early training.

Empirical studies further show that routing granularity influence the emergence of expert functions. Sequence-level gating (i.e., one routing decision per input) typically groups experts by topic or discourse structure, whereas token-level gating yields more fine-grained specialization that frequently aligns with syntactic categories such as nouns or verbs~\cite{2402.13089}.
\textbf{Emergent linguistic structure in expert assignments.}  
Probing experiments~\cite{2412.16971} reveal that MoE layers implicitly cluster inputs by part-of-speech and morphological role, even without explicit supervision. Let $S_i = \{x_t \mid g_i(x_t) > 0\}$ be the token set for expert $i$; statistical analysis shows that $S_i$ has high mutual information with specific POS categories:
\begin{equation}
    I(\text{POS}; \text{Expert}_i) = H(\text{POS}) - H(\text{POS} \mid S_i).
\end{equation}
This implies that specialization arises from the interplay between architecture and training dynamics, not solely from design choices. In practice, this interpretability enhances modular debugging, domain adaptation, and controllable generation.

\textbf{Adaptive expert selection.}  
In adaptive MoE architectures, the gating function $g(x)$ is augmented with input-dependent capacity control, enabling dynamic scaling of expert participation. For input $x$, the number of activated experts $k(x)$ is computed as:
\begin{equation}
    k(x) = \min\left(K_{\max}, \left\lfloor \tau \cdot \|x\| \right\rfloor \right),
\end{equation}
where $\tau$ is a learnable or fixed scaling coefficient and $\|x\|$ denotes input complexity (e.g., norm, entropy, or proxy task difficulty)~\cite{zoph2022st}.

This enables the router to allocate deeper or wider sub-networks for semantically rich or ambiguous inputs, while routes simpler tokens to lightweight modules. In multi-modal settings, adaptive expert counts have shown measurable gains in sample efficiency and robustness under domain shift, particularly when expert allocation aligns with modality boundaries (e.g., image tokens vs. text tokens). Unlike static MoEs, these models respond to context shifts by reshaping expert usage dynamically, without retraining.

In continual learning, dynamic routing helps preserve old knowledge while incorporating new, mitigating catastrophic forgetting~\cite{mcclelland2020placing}. It also improves robustness to distribution shifts, adapting expert usage to evolving input domains~\cite{bengio2009curriculum}.

\section{Meta-Learning and Knowledge Transfer in MoE}
While the architectural foundations we've explored provide the building blocks for effective MoE systems, and the advanced variants of MoE model expertizes in dealing with a variety of critical challenges,  a key question remains: how can these models learn to learn and transfer knowledge across different domains? This section examines meta-learning's development with MoE capabilities, which enables rapid adaptation and efficient knowledge transfer.

\subsection{Meta-Learning Framework Design}

Meta-learning enhances MoE systems by enabling rapid generalization across diverse tasks without retraining from scratch~\cite{bischof2022mixture, liu2024meta}. Instead of learning task-specific routing independently, meta-MoE architectures optimize a routing policy $\theta$ across a distribution of tasks $\mathcal{T}$, enabling fast adaptation to unseen tasks with limited support data:
\begin{equation}
\theta_{\mathcal{T}_\text{new}} = \theta - \eta \nabla_\theta \mathcal{L}_{\text{support}}(\theta),
\end{equation}
where $\mathcal{L}_{\text{support}}$ denotes the support loss induced by the sparse expert outputs.

\textbf{Hierarchical meta-learning with MixER.}  
To handle nested dynamical systems, the MixER model (Mixture of Expert Reconstructors) introduced in ~\cite{2502.05335} augments classical MoE layers by feeding an additional context vector $\xi$ alongside the input $x$ into the router. As shown in \textbf{Figure~\ref{fig:MixER}}, MixER bypasses softmax gating and instead performs discrete expert selection via a K-means-inspired objective:
\begin{equation}
z(x, \xi) = \arg\min_j \left\| f_\theta(x, \xi) - \mu_j \right\|^2,
\end{equation}
where $f_\theta$ maps $(x, \xi)$ to a latent space, and $\mu_j$ denotes the prototype of expert $j$.

This configuration enables top-1 routing with interpretable cluster assignments and avoids the overhead of differentiable soft selection. It demonstrates strong performance on sparse reconstruction tasks, such as parametric ODE systems. However, when contextual hierarchies are weak or missing, specialization degrades due to overlapping token-to-expert assignments.

\textbf{Meta-distillation for domain adaptation.}  
To address domain shift, the Meta-DMoE framework proposed in ~\cite{2210.03885} formulates test-time adaptation as a meta-distillation problem. A set of domain-specific experts $\{E_i\}$ are pre-trained on disjoint source domains $\{\mathcal{D}_i\}$, and their predictions are aggregated through a transformer-based aggregator $\mathcal{A}$ to supervise a lightweight student model $S$:
\begin{equation}
\mathcal{L}_{\text{meta}} = \text{KL}\left(S(x) \,\|\, \mathcal{A}\left(E_1(x), \dots, E_N(x)\right)\right).
\end{equation}

The aggregator learns to combine the expert outputs based on inter-domain dependencies, while meta-optimization ensures that the knowledge is transferable to unseen target domains. This approach improves generalization in scenarios where domain labels are unavailable or where training a unified model on heterogeneous domains is suboptimal.

\subsection{Knowledge Transfer Mechanisms}
\textbf{Sparse-to-Dense knowledge integration.} 
Transferring knowledge from sparse MoE models to dense architectures presents significant challenges, prompting researchers to propose several effective solutions. Inspired by human pedagogical frameworks, a multi-teacher distillation strategy has been proposed, wherein multiple expert models collaboratively supervise a single student model. This setup enables the student to integrate diverse knowledge sources, improving generalization across tasks~\cite{2201.10890}. Their framework tackles some real problems with sparse MoE models, which tend to overfit, are tricky, and often don't play nice with existing hardware. The core idea is straightforward: Instead of trying to compress everything into one dense model directly, you let multiple experts teach a single student model. This knowledge integration approach sidesteps many of the headaches that come with sparse MoE deployment.

The proposed framework encompasses knowledge gathering and knowledge distillation phases. Four distinct knowledge gathering methods are investigated: summation, averaging, Top-K Knowledge Gathering (Top-KG), and Singular Value Decomposition Knowledge Gathering (SVD-KG). The dense student model (OneS) preserves 61.7\% of MoE benefits on ImageNet, achieving 78.4\% top-1 accuracy with only 15M parameters. On natural language processing datasets, OneS obtains 88.2\% of MoE benefits while outperforming baselines by 51.7\%. The approach achieves 3.7× inference speedup compared to MoE counterparts due to reduced computation and hardware-friendly architecture.

\textbf{Mutual distillation among experts.}
A known limitation of MoE architectures is their narrow learning scope: individual experts often struggle to generalize due to restricted exposure to diverse training samples. This challenge has been addressed by incorporating mutual distillation mechanisms. The MoDE (Mixture-of-Distilled-Expert) framework~\cite{2402.00893} introduces moderate mutual distillation among experts, facilitating knowledge sharing and enhancing task awareness.

The MoDE framework addresses the fundamental issue where gate routing mechanisms restrict experts to limited sample exposure, thereby constraining generalization ability improvements. Through moderate mutual distillation, each expert acquires features learned by other experts, gaining more accurate perceptions of their originally allocated sub-tasks. Extensive experiments across tabular, natural language processing, and computer vision datasets demonstrate MoDE's effectiveness, universality, and robustness. The approach is validated through innovative "expert probing" studies. These validations demonstrate that moderate knowledge distillation improves individual expert performance, thus enhanced the overall MoE performance.

\subsection{System Platform Support}

The deployment of meta-learning-based MoE systems demands not only model-level optimization but also end-to-end systems engineering support. AwesomeMeta+~\cite{2304.12921} addresses this need by introducing a standardized prototyping platform that encapsulates core meta-learning components as reusable and configurable modules. This modular design abstracts recurrent patterns such as task conditioning, gradient aggregation, and adaptation loops into composable units, simplifying integration into broader training and inference pipelines.

A key contribution of AwesomeMeta+ lies in its ability to translate theoretical meta-learning constructs into deployable MoE workflows. Traditional meta-learning implementations tend to be highly task-specific, limiting their extensibility and reproducibility. AwesomeMeta+ mitigates this through a layered architecture comprising: (i) a declarative model interface that maps task descriptors to expert selectors, (ii) a scheduler that optimizes expert instantiation under resource constraints, and (iii) an evaluation monitor that tracks transferability metrics such as few-shot accuracy or expert stability across tasks.

To assess platform effectiveness, both automated benchmarking and user studies were conducted. Feedback from over 50 researchers indicated that the framework improves understanding of meta-adaptation logic and accelerates system assembly. Moreover, measured performance across meta-dataset benchmarks confirms that the platform incurs negligible overhead while enabling consistent deployment of otherwise fragmented designs. By bridging theoretical innovation and practical engineering, AwesomeMeta+ demonstrates that platform-level standardization is essential for scalable, adaptable, and maintainable MoE-based meta-learning systems.

\begin{figure*}[t]
\centering
\includegraphics[width=\linewidth,height=0.6\textheight,keepaspectratio]{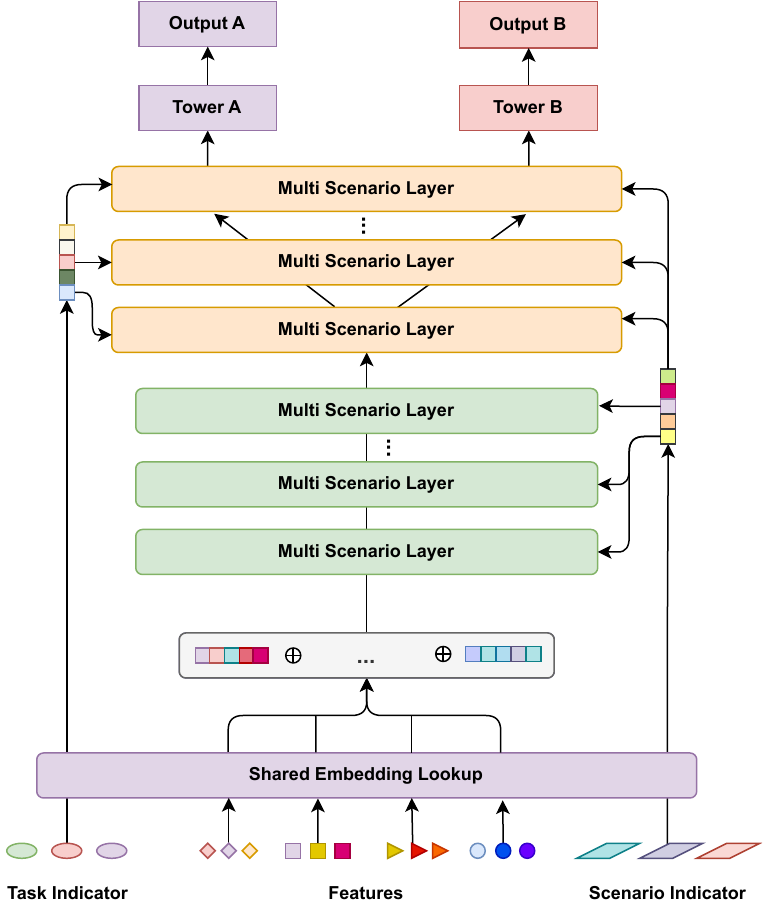}
\caption{\textbf{The Architecture of AESM$^2$ for Multi-Task Learning.} The framework is composed of major modules including Shared Embedding Layer used to cast the raw categorical and numerical features into continuous embeddings, and Multi-Scenario Layer used for expert selection, as well as Multi-Task Layer for multi task learning.}
\label{fig:AESM2}
\end{figure*}
\section{MIXTURE OF EXPERTS APPLICATIONS AND DOMAIN-SPECIFIC MODELS}
Having reviewed the theoretical framework and learning mechanisms of MoE, we now examine the applications of it in real world. In this section, we will examine the adoption of MoE across recommendation systems, search, computer vision, NLP, healthcare, etc.  These models are transforming diverse fields by taking advantage of their specialized expertise to solve  real-world challenges.

\subsection{Recommendation Systems and Search}

In large-scale recommendation and search applications, MoE architectures are increasingly employed to handle the inherent complexity of multi-domain and multi-task personalization~\cite{2404.18465, hou2022towards}. Conventional dense models often struggle to balance domain-specific signal extraction and shared knowledge transfer, especially under rapidly changing user contexts, traffic distributions, and item lifecycles.

To address these limitations, M3oE~\cite{2404.18465} proposes a modular MoE framework that jointly models domain- and task-level heterogeneity. It deploys parallel expert modules to capture shared user preferences, domain-specific behavior, and task-specific patterns, which are fused through a hierarchical gating mechanism. By integrating AutoML-based structure search, M3oE can adapt its expert composition over time, improving robustness and scalability under production workloads.

As illustrated in \textbf{Figure~\ref{fig:AESM2}}, AESM2 follows a similar design philosophy but targets scenario-aware recommendation. It begins with a shared embedding layer over input features, scenario indicators, and task indicators, followed by stacked multi-scenario layers that refine contextual representations across traffic segments. Scenario-informed representations are then routed into task-specific towers through a hierarchical routing mechanism that selects experts at both scenario and task levels. This design enables controlled knowledge sharing while allowing specialization when traffic patterns diverge. Empirical results show that AESM2 improves retrieval quality and training stability under dynamic traffic shifts, outperforming static multi-gate MoE baselines.

\begin{figure*}[t]
\centering
\includegraphics[width=\linewidth]{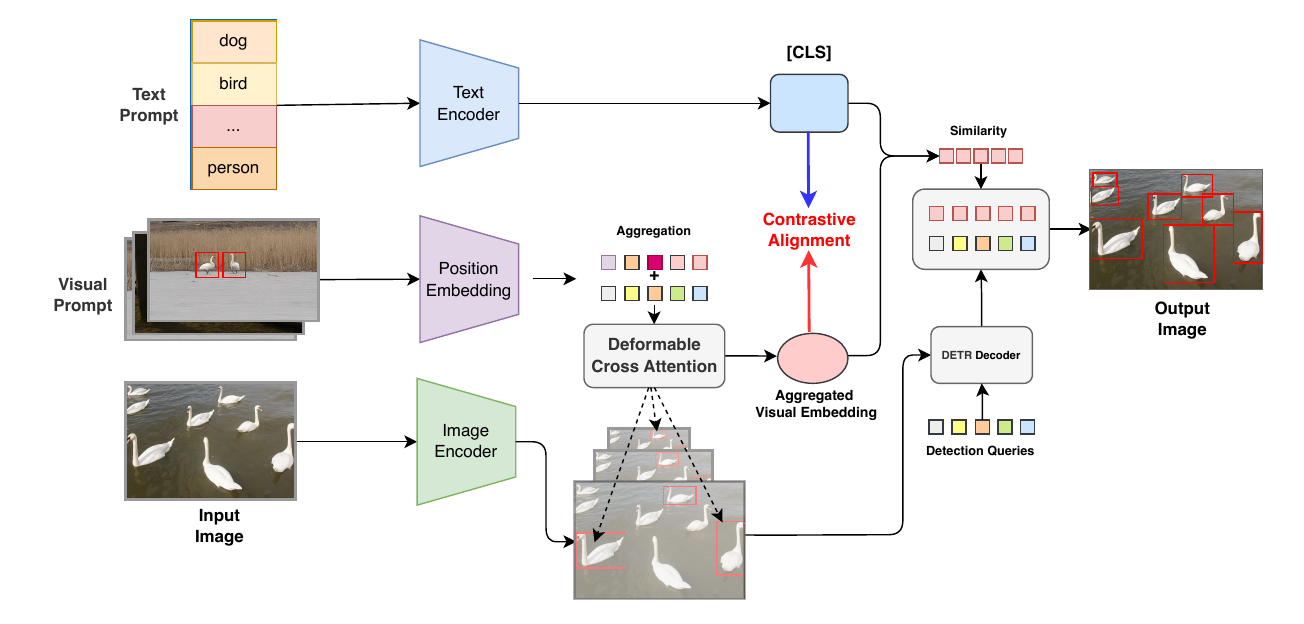}
\caption{\textbf{Architectural overview of the T-Rex2 framework.} The model adopts DETR-based design methodology for end-to-end object detection~\cite{carion2020end}. Visual and textual prompts are processed via deformable cross-attention~\cite{zhu2021deformable}  mechanisms and CLIP~\cite{radford2021learning} text encoding respectively, with multimodal alignment achieved through contrastive learning strategies.}
\label{fig:T-Rex2}
\end{figure*}

\subsection{Multimodal and Multitask Learning}

MoE architectures have proven effective in addressing the inherent complexity of multimodal and multitask learning, where diverse input modalities and objective functions must be handled simultaneously. Recent advancements in models such as MoVA~\cite{zong2024mova}, DeepSeek-VL2~\cite{wu2024deepseek}, Omni-SMoLA~\cite{2312.00968}, T-REX2~\cite{jiang2024t}, MoME~\cite{shen2024mome}, and MoTE~\cite{wang2025mote} collectively reflect a growing consensus that fine-tuning alone is insufficient for achieving robust generalization across heterogeneous tasks and data types.

Omni-SMoLA addresses a core limitation in large multimodal models: task interference from shared representations. By integrating low-rank expert modules specialized for different modalities and tasks, it enables modular specialization while preserving generalist capabilities. Empirical studies show that compared to standard LMM fine-tuning, SMoLA’s sparse expert routing leads to more stable convergence and better performance across diverse tasks, suggesting that architectural modularity is key to scalable multimodal generalization~\cite{2312.00968}.

\textbf{Multimodal specialization via hybrid prompts.}
T-REX2 further advances open-set object detection by exploiting the complementary strengths of textual and visual prompts. As depicted in \textbf{Figure~\ref{fig:T-Rex2}}, the model incorporates dual encoders, with one responsible for processing abstract text categories (e.g., “dog”, “bird”), and the other extracting instance-level features from visual exemplars. These streams are fused through a deformable cross-attention module that integrates position-aware visual embeddings with context from prompts. A contrastive alignment loss links the aggregated visual embedding with the [CLS] representation from the text encoder, enforcing semantic consistency across modalities.

This hybrid prompting design enables robust generalization to unseen classes, where text offers conceptual grounding and visual prompts provide fine-grained instance cues. Unlike single-modal methods, T-REX2 flexibly accommodates different prompt combinations and achieves high zero-shot detection accuracy across diverse domains~\cite{jiang2024t}.

\textbf{Expert reuse via hypernet-based modulation.}
HyperMoE addresses a central limitation in sparse expert models: the underutilization of unselected experts during forward inference. In standard MoE architectures, only the top-$k$ experts are activated per input, leaving the rest unused. This underutilization of capacity can impair generalization, especially in multitask or low-resource scenarios.

To overcome this, HyperMoE introduces a hypernetwork that leverages the hidden states of inactive experts to generate lightweight modulation signals. These signals are injected into the output paths of the active experts, enabling implicit expert collaboration without evaluating the full expert set. This mechanism retains routing sparsity while enriching the active computation with global knowledge from the broader model.

Recent research attention in multimodal MoE continues to refine interaction-aware routing and modality-specific expert specialization to handle richer and more heterogeneous data: I2MoE~\cite{xin2025i2moe} proposes an interpretable interaction-aware routing framework where distinct experts are trained to capture diverse multimodal interaction patterns, providing local and global interpretability of expert contributions. This strategy enhances modality fusion by explicitly modeling how different sources of information interact during routing. Uni3D-MoE~\cite{zhang2025uni3dmoe} targets 3D multimodal scene understanding by integrating modalities such as multi-view RGB, depth maps, and point clouds. Its learned routing mechanism selects experts based on modality preferences and task-specific context, enabling flexible collaboration for tasks like 3D object recognition and spatial understanding. SMAR~\cite{xia2025smar} introduces a soft modality-aware routing strategy for multimodal MoE models that regularizes routing probabilities across modalities, preserving language capabilities when adapting pretrained MoE models to multimodal objectives. This approach balances modality separation and generalization without altering model architecture or requiring additional modalities during pretraining. MoDES~\cite{huang2025modes} focuses on efficient multimodal MoE inference, proposing a training-free dynamic expert skipping mechanism tailored for heterogeneous contributions across experts and layers in multimodal large language models, decreasing computation while preserving accuracy.

\begin{figure*}[t]
\centering
\includegraphics[width=0.95\linewidth]{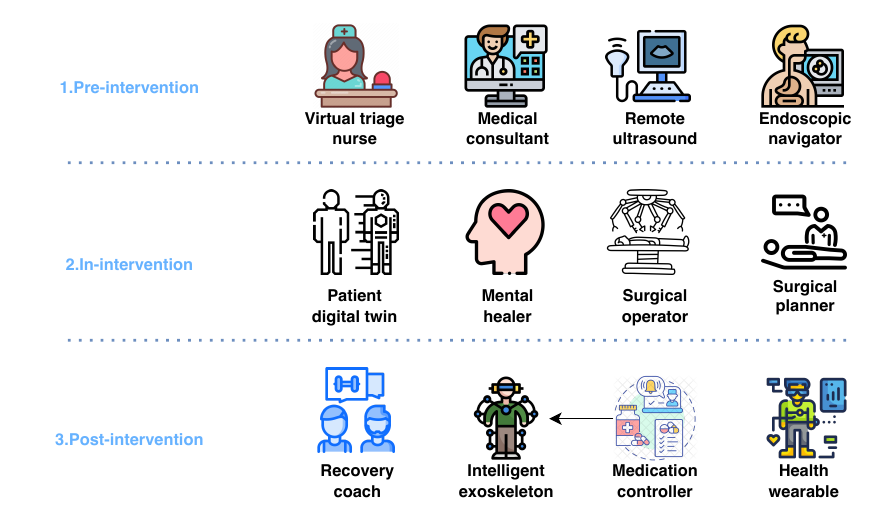}
\caption{\textbf{Applications of embedded AI in healthcare.} As the figure shows, embodied AI has been used in health care scenarios including but not limited to: pre-intervention (virtual triage nurse, medical consultant, remote ultrasound, endoscopic navigator), in-intervention (patient digital twin, mental healer, surgical operator, surgical planner), and post-intervention (recovery coach, intelligent exoskeleton, medication controller, health wearable.}
\label{fig:health_care}
\end{figure*}

Empirical results show that such modulation enhances downstream performance, particularly on tasks with limited data, and improves expert diversity without increasing computational cost. The approach exemplifies how structural innovations can balance expressiveness and efficiency in scalable multitask systems.

\subsection{Healthcare and Life Sciences}
MoE architectures are increasingly applied in healthcare, addressing key challenges in patient care, clinical decision-making, and system efficiency. Representative models include Med-MoE~\cite{pieri2024bimedix}, BiMediX~\cite{pieri2024bimedix}, and LoRA-based medical MoEs~\cite{chen2024low}. Given the safety-critical nature of medical applications, these systems emphasize accuracy, modularity, and interpretability, aiming to support diagnostic reasoning while remaining compatible with clinical constraints.

A significant frontier is Embodied Intelligence in healthcare, where robotic systems assist with elderly care, rehabilitation, and clinical procedures. As shown in \textbf{Figure~\ref{fig:health_care}}, embodied agents leverage perception, actuation, planning, and memory across tasks ranging from bedside assistance to surgical support~\cite{2501.07468}. However, barriers such as limited integration into existing workflows, simulation-to-reality gaps, and the absence of standardized evaluation benchmarks persist, limiting widespread deployment.

To address the scarcity of medical data, Syn-Mediverse~\cite{2308.03193} introduces a large-scale synthetic dataset comprising over 48,000 hyper-realistic images and 1.5 million annotations across five vision tasks, enabling robust visual perception in complex healthcare environments. Building on this, AT-MoE~\cite{2410.10896} enhances interpretability and specialization through LoRA-tuned expert layers and grouped adaptive routing, which dynamically fuses task-relevant modules to support controllable, transparent decision-making. These advancements highlight the need to balance technological innovation with safety, interpretability, and ethical considerations in life-critical medical applications.

Recent work after May~2025 further explores MoE for robust multimodal clinical prediction and anatomically informed medical imaging. MoE-Health~\cite{wang2025moe_health} introduces a dynamic gating Mixture-of-Experts framework tailored for healthcare prediction with heterogeneous and incomplete input modalities (electronic health records, clinical notes, and medical imaging). MoE-Health dynamically selects and integrates specialized expert modules based on available data, improving performance on tasks such as in-hospital mortality, length-of-stay estimation, and readmission prediction under real-world variability in data availability. Such robustness is crucial for clinical deployment where complete multimodal inputs are rarely guaranteed. In medical imaging, MedMoE~\cite{chopra2025medmoe} proposes modality-specialized expert branches within a vision-language MoE framework that adapts feature extraction to diagnostic context, enhancing alignment between imaging and textual findings across diverse clinical benchmarks. Additionally, REN (Regional Expert Networks)~\cite{peltekian2025ren} leverages anatomical priors to train region-specific expert modules for interstitial lung disease diagnosis, demonstrating improved classification performance and interpretability by coupling expert outputs with anatomical structure information. A related trend involves specialized MoE strategies that improve robustness and generalization in medical multimodal learning. For example, adaptive expert grouping mechanisms—designed to exploit collaborative tendencies among experts—have been introduced to reduce routing overhead while preserving generalization benefits in medical contexts~\cite{ICLR2026FineGrainedMedicalMoE}. 

A significant frontier remains embodied intelligence in healthcare, where robotic and assistive systems integrate perception, planning, and action in physical settings. As shown in \textbf{Figure~\ref{fig:health_care}}, embodied agents leverage cross-modal perception and learned policies across tasks ranging from bedside assistance to rehabilitation. However, barriers such as integration into clinical workflows, simulation-to-reality gaps, and the absence of standardized evaluation benchmarks persist, limiting widespread deployment. Efforts that tightly couple MoE decision mechanisms with domain constraints and real-world validation pipelines may help bridge this gap.

\subsection{Computer Vision and Image Processing}
Modern computer vision has progressed from CNN-based pipelines to Transformer and diffusion-based architectures. MoE designs are increasingly integrated into this evolution to address the growing complexity of tasks such as object detection, image classification, and scene understanding. Representative applications include AdaMV-MoE~\cite{chen2023adamv}, GNT-MOVE~\cite{Cong_2023_ICCV}, and recent efforts in image classification using expert-based decomposition~\cite{videau2024mixture}.

In object detection, the Mixture of Calibrated Experts (MoCaE)~\cite{2309.14976} framework introduces a principled approach to aggregating predictions from multiple detectors. Traditional ensemble-based detectors suffer from miscalibrated confidence outputs, where dominant experts may overwhelm the consensus even in uncertain regions. MoCaE addresses this by calibrating each expert’s output based on empirical performance, leading to more reliable prediction fusion. Evaluated on COCO and related benchmarks, MoCaE achieves up to +2.5 AP improvement, establishing new performance baselines.

The issue of effective expert specialization has also been tackled through refined gating architectures and regularization~\cite{2302.14703}. Early MoE models often failed to disentangle task-relevant features, resulting in homogeneous expert behavior. To address this, attention-like gates combined with entropy-minimizing regularizers enable low-overlap, semantically aligned expert selection. Experiments across classification datasets (e.g., MNIST \cite{lecun2002gradient}, CIFAR \cite{krizhevsky2009learning}, FashionMNIST \cite{xiao2017fashion}) demonstrate that this gating scheme improves both accuracy and interpretability of expert routing.

Deeper variants such as the Deep Mixture of Experts model~\cite{1312.4314} propose hierarchical expert compositions with stacked routing layers. This architecture separates “where” experts in early spatial layers from “what” experts in deeper semantic stages, allowing task-conditional specialization without inflating parameter count. These designs reflect a broader trend in visual MoEs: balancing model sparsity with fine-grained expressiveness.

By enabling modular learning over hierarchical visual signals, MoE frameworks have demonstrated efficacy in managing multi-scale feature hierarchies, adaptive capacity allocation, and task-specific specialization within scalable visual pipelines.

\subsection{Natural Language Processing and Large Language Models}
Among all domains, NLP and LLMs have witnessed the most impactful and widespread adoption of MoE architectures. The fundamental motivation is clear: enabling scalable capacity without proportional increases in inference or training cost~\cite{artetxe2022efficient}. This has led to numerous architectural advances that have redefined efficiency paradigms in the tech industry and open-source communities alike.

A key limitation of early MoE systems is their incompatibility with parameter-efficient fine-tuning (PEFT), due to the need to store and update a full set of experts. To overcome this, a recent study proposes an extremely parameter-efficient MoE framework~\cite{2309.05444}, which replaces dense expert networks with lightweight modules, achieving performance comparable to full fine-tuning while updating less than 1\% of parameters in an 11B-scale model. This makes fine-tuning feasible even in constrained environments, without sacrificing task-specific adaptability.

Complementing this, flexible composition frameworks for domain adaptation have emerged. A toolkit for building low-cost Mixture-of-Domain-Experts (MoDE)~\cite{2408.17280} enables combining trained adapters or full models into expert pools tailored for specific domains. This method supports modular domain composition without retraining from scratch, offering practical deployment value. The framework includes guidance for optimal configuration and has demonstrated efficacy in multi-domain scenarios under constrained compute.

On the theoretical side, recent work on MoE hypothesis construction~\cite{2406.17150} deepens our understanding of their representational properties. Unlike Bayesian ensembles, which aggregate across uncertainty distributions, MoEs select hypotheses via discrete routing. This mechanism effectively enables abductive reasoning over the hypothesis space. Under mild assumptions, it is shown that MoE models exhibit higher functional capacity and can outperform Bayesian alternatives in certain regimes, even when relying on weaker inductive priors.

Recent work continues advancing MoE design for NLP. MoxE introduces extended LSTM-based MoE with entropy-aware routing to balance expert utilization and improve efficiency in language modeling, showing improved scalability with rare token handling~\cite{moxe2025}. Another direction is lightweight low-rank adaptation Mixture-of-Experts (L-MoE), which unifies MoE and LoRA in an end-to-end trainable framework where task-specialized low-rank expert adapters are dynamically composed via differentiable routing, improving parameter efficiency and dynamic skill composition~\cite{lmoe2025}.

These findings collectively underscore the strength of MoE models in language representation tasks, combining modularity, adaptability, and efficiency. As a result, MoE architectures continue to play a central role in building scalable LLM systems that are both performant and cost-effective.

\subsection{Methodological Innovation and Theoretical Foundations}
The success of MoE architectures is grounded in a growing body of theoretical insights and methodological advancements that address scalability, convergence properties, expert diversity, and model integration.

Scaling behavior under memory constraints has been systematically explored through unified scaling laws for dense and sparse models~\cite{2502.05172}. By incorporating active parameter counts, dataset sizes, and expert configurations, the study demonstrates that MoE models can surpass dense models in memory efficiency, contrary to earlier assumptions. Empirical evaluations over parameter regimes up to 5B validate this advantage, offering guidance for efficient training in constrained environments.

The convergence behavior of MoE parameter estimation has been formalized in recent work on Gaussian-gated models~\cite{2305.07572}. By analyzing maximum likelihood estimation (MLE) under gating covariates, the study introduces Voronoi-based loss functions to characterize non-uniform convergence rates. The analysis reveals how different configurations of location parameters lead to distinct solution spaces governed by polynomial systems, providing a deeper understanding of optimization dynamics in sparse settings.

Recent work further strengthens the theoretical foundation of MoE systems by reframing routing and expert interaction as structured optimization problems. Omi et al.~\cite{omi2025similarity_router} introduce a similarity-preserving load-balancing objective that enforces consistent expert assignment for related inputs, yielding provable reductions in expert collapse and variance over long training horizons. Complementarily, \textsc{MaxScore} routing~\cite{dong2025maxscore} formulates expert selection as a minimum-cost maximum-flow problem, providing an explicit optimization framework that trades off expert capacity, token assignment, and communication cost. These formulations move beyond heuristic regularization and offer principled guarantees on stability and efficiency.

Expert diversity and inter-expert knowledge transfer are addressed through several architectural strategies. Orthogonal MoE (OMoE)~\cite{2501.10062} enforces expert specialization via Gram-Schmidt orthogonalization of expert weights, while mutual distillation in MoDE and related frameworks~\cite{2402.00893, huang2024improving} encourages knowledge sharing through pairwise KL divergence. These mechanisms enhance feature coverage and mitigate redundancy. Nexus~\cite{2408.15901} introduces adaptive routing and parameter reuse, enables the conversion of dense models into sparse expert systems without retraining from scratch.

To improve generalization and model merging, hypernetwork-based strategies have emerged. HMoE~\cite{2211.08253} leverages hypernetworks to generate expert parameters dynamically, supporting low-dimensional alignment across domains. Concurrently, parameter merging frameworks~\cite{2502.00997} address conflicts in heterogeneous expert integration, introducing alignment and reparameterization strategies that preserve performance while minimizing interference.

Collectively, these developments reinforce the theoretical robustness of MoE models and expand their applicability to complex training regimes, where capacity, adaptability, and diversity must be simultaneously managed.

\begin{figure*}[t]
\centering
\includegraphics[width=\linewidth]{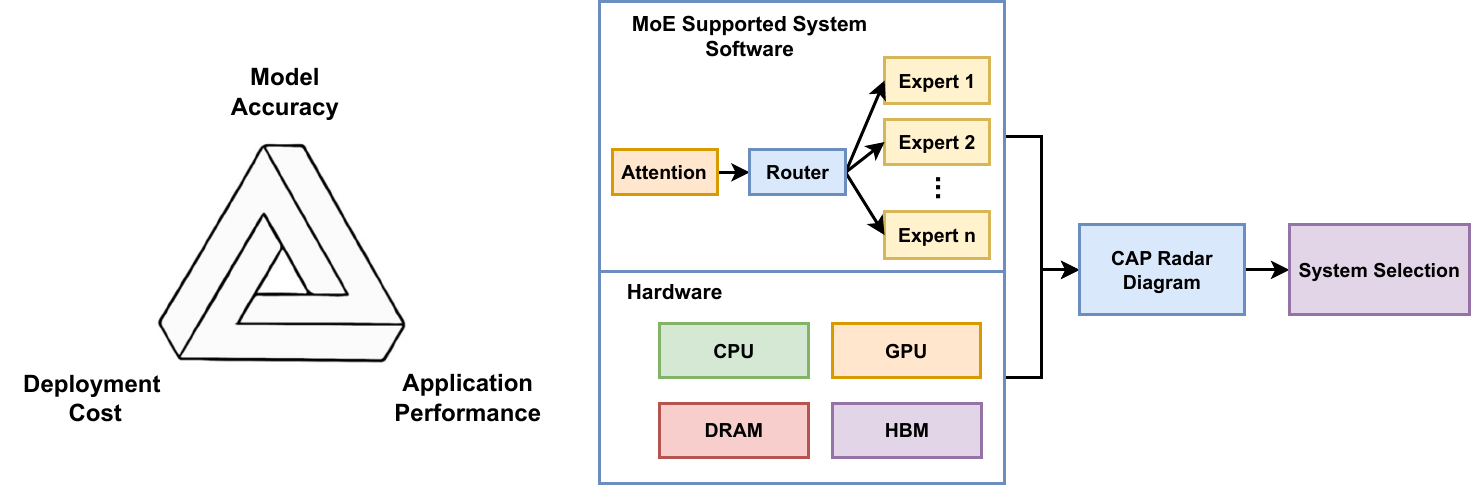}
\caption{\textbf{Framework illustration of MoE-CAP methodology.} Left: The triangular relationship demonstrates the balance among deployment cost, model accuracy, and application performance in MoE system design. Right: The MoE-CAP framework employs novel sparsity-aware evaluation metrics and CAP radar visualization to provide comprehensive assessment capabilities for MoE architectures, facilitating informed decisions regarding system architecture and hardware configuration selection.}
\label{fig:Mo-CAP}
\end{figure*}

\section{Evaluations, Challenges and Future Directions}
As MoE models gain wider adoption in real-world applications, critical questions about their evaluation and limitations are becoming increasingly important. Although traditional benchmarks used to evaluate LLMs, such as LLM-Perf Leaderboard~\cite{llm-perf-leaderboard}, Mlperf inference benchmark~\cite{reddi2020mlperf} and  Mmbench~\cite{xu2023mmbench}, are very popular and powerful tools, they are not well-suited in evaluating MoE models. A methodology to evaluate MoE models and provide guidance in the design choice, is one of the urgent needs. This section examines the development of methods for evaluating MoE performance, the remaining challenges, and the future directions for research and development.
\subsection{Evaluation Framework and Methodology}
\textbf{Theoretical foundation and evaluation principles.} 
The evaluation of MoE architectures presents unique challenges that stem from their fundamental design principles. Unlike traditional dense models, MoE systems are conditional computation, which only activates specific expert subsets for each input, and require special attention. In these architectures, there are many factors that affect the system's performance, including the interplay between expert specialization, routing. This evaluation frameworks need to be aware of those factors, and accurately capture their effects.

To be more specific, MoE's divide-and-conquer principle is more complex than dense models, as the problem space partition's effectiveness directly impacts model performance. Apart from evaluation of the final output, evaluation frameworks also needs to assess the intermediate processes of expert assignment, load balancing, and knowledge distribution across the expert ensemble, which is a end to end process.

\textbf{Standardized evaluation framework.} There is a critical need for comprehensive benchmarking platforms, since advancing MoE research has encountered a blocker: the lack of standard measure, and guidance. Recently researchers have been proposing studies on new benchmarks, some notable ones are Mixtral 8x7B~\cite{mlcommons2024mixtral} and LibMoE~\cite{nguyen2024libmoe}. LibMoE is especially noteworthy since it presents a modular framework, covering streamlined research, training, and evaluation steps of MoE algorithms, basically the full lifecycle.  

To validate the framework's benchmarking capabilities, researchers executed experiments and were astonished by the findings: by executing systematic evaluation over the five state-of-the-art MoE algorithms across three different large language models and eleven datasets under zero-shot settings~\cite{nguyen2024libmoe}, LibMoE have yielded critical insights. One notable finding is: although the evaluated MoE algorithms were developed on different task purposes, all of them achieved similar average performance across a wide range of tasks. This finding suggests that the choice of the MoE algorithm may be less critical than previously assumed.

\textbf{System-level multi-dimensional evaluation.}
The deployment of MoE architectures in real-world systems requires evaluation beyond model accuracy. As shown in \textbf{Figure~\ref{fig:Mo-CAP}}, the MoE-CAP framework~\cite{fu2024moe} introduces a triadic evaluation paradigm encompassing \emph{model accuracy}, \emph{application performance}, and \emph{deployment cost}. This CAP triangle highlights the inherent trade-offs among the three, where optimizing two often comes at the expense of the third.

To operationalize this evaluation, MoE-CAP integrates software- and hardware-level profiling. The software stack includes attention mechanisms, routers, and expert networks, while the hardware layer incorporates compute and memory components (e.g., CPU, GPU, DRAM, HBM). By analyzing the interaction between routing sparsity and hardware utilization, the framework generates CAP radar plots to support system comparison under constraints such as latency or budget. This enables structured, deployment-aware architecture selection aligned with practical feasibility rather than accuracy alone.

\textbf{Specific evaluation method examples.} The Mixture of Calibrated Experts (MoCaE) framework~\cite{2309.14976} offers a practical enhancement to MoE evaluation by addressing a persistent issue: \emph{prediction miscalibration}. In conventional MoE models, expert outputs are often fused based on confidence scores that do not reliably reflect true accuracy, resulting in suboptimal predictions dominated by overconfident experts.

MoCaE mitigates this by introducing calibration procedures prior to expert output aggregation. Instead of directly averaging raw predictions, each expert’s output is first adjusted to better reflect its empirical reliability. This calibrated fusion improves ensemble robustness and reduces overfitting to dominant experts. Empirical results on the COCO benchmark demonstrate performance gains of up to 2.5 AP, establishing MoCaE as a new state-of-the-art approach across multiple object detection tasks.

\textbf{Expert diversity and representation learning challenges.} A fundamental limitation in MoE architectures lies in the lack of expert specialization. Empirical studies have shown that experts often converge to nearly identical representations, with similarity scores exceeding 99\% across diverse inputs~\cite{2310.09762}. Notably, this phenomenon is not confined to underperforming or poorly regularized models, but also emerges in high-performing configurations, suggesting it is a systemic issue.

Such representational homogeneity directly undermines the divide-and-conquer principle that motivates the MoE paradigm. Without sufficient diversity, experts fail to develop complementary capabilities, leading to inefficient parameter utilization and degraded task generalization. This challenge highlights the need for architectural and training mechanisms that explicitly promote expert differentiation.

\textbf{Architectural design and integration challenges.} The integration of shared layers within MoE architectures has been observed to degrade performance in certain configurations~\cite{2405.11530}. One plausible explanation is that experts learn redundant or conflicting representations when exposed to identical shared features, thereby reducing specialization and increasing interference. This suggests that naïve parameter sharing may hinder the decomposition of tasks and limit the expressiveness of individual experts.

In addition, dynamic expert expansion in incremental learning settings poses nontrivial challenges. When new experts are added post hoc~\cite{1511.06072}, the system must resolve conflicts arising from inconsistent outputs among parallel experts. These inconsistencies can destabilize training or lead to suboptimal predictions. Addressing this requires the development of conflict-aware routing or mediation strategies that ensure coherence across dynamically evolving expert pools.

\textbf{Routing mechanisms and specialization challenges.} The necessity and efficacy of learned routing mechanisms in MoE systems remain an open and debated question. Empirical studies have shown that frozen, randomly initialized routers can perform on par with learned routing strategies across several benchmarks~\cite{2402.13089}. These findings challenge the commonly held assumption that adaptive routing is critical for MoE performance and suggest that increased routing complexity may not always yield proportional benefits. This raises important questions regarding the trade-off between routing expressiveness and architectural simplicity, particularly in low-resource or latency-constrained environments.

\textbf{Challenges in Theoretical Grounding of MoE Architectures.}  
Despite the empirical success of MoE architectures, particularly within natural language processing tasks, their theoretical foundations remain underdeveloped~\cite{chen2022towards, mu2025comprehensive}. Most existing designs rely on experimental heuristics rather than principled models, leaving unresolved key questions about the relationship between expert diversity, specialization dynamics, and system-level generalization.

A quantitative framework linking expert diversity to generalization and modular efficiency would guide principled MoE design, setting expert count, garsity and gating for any task or data. Formal analysis of routing and selection, grounded in information or learning theory, can replace costly empirical tuning and shrink the design space.

\textbf{Technical method innovation.}
Efforts to improve expert diversity and routing precision in MoE architectures have led to a range of architectural innovations. Methods such as DeepSeekMoE~\cite{dai2024deepseekmoe}, TableMoE~\cite{zhang2025tablemoe}, and Pre-gated MoE~\cite{hwang2024pre} introduce input-aware expert allocation and gating preconditioning to enhance functional specialization. Orthogonal Mixture-of-Experts (OMoE)~\cite{2501.10062}, in particular, applies orthogonality constraints on expert weights to reduce representational redundancy, thereby encouraging more disentangled expert behavior.

While such structural designs improve modularity, their potential can be further enhanced by incorporating feedback-based optimization. For instance, reinforcement learning techniques, including Reinforcement Learning from Human Feedback (RLHF), can guide expert selection and adjust routing policies using reward signals aligned with human preferences. These observations indicate that hybrid approaches combining architectural regularization with adaptive learning strategies hold significant promise for building more robust and generalizable MoE systems.

\section{Conclusion}
This survey provides a comprehensive overview of recent advances in Mixture of Experts architectures. We first traced the evolution of MoE from its theoretical origins to large-scale implementations, followed by an in-depth analysis of core architectural components and design principles. We then examined advanced variants, meta-learning frameworks, and knowledge transfer mechanisms, as well as domain-specific applications across industry and academia. Furthermore, we discussed current evaluation methodologies, highlighted major challenges such as routing stability and expert specialization, and identified promising directions for future research. We hope this work serves as a valuable resource for researchers and practitioners, and contributes to the continued development of scalable and efficient MoE-based systems.

\bibliographystyle{plain}
\bibliography{cite}

\end{document}